\documentclass[runningheads]{llncs}
\usepackage{graphicx}
\usepackage{amsmath,amssymb} 
\usepackage{color}
\usepackage[width=122mm,left=12mm,paperwidth=146mm,height=193mm,top=12mm,paperheight=217mm]{geometry}

\usepackage{epsfig}
\usepackage{graphicx}
\usepackage{wrapfig}

\usepackage{amsmath, amsthm, amssymb}
\usepackage{textcomp}
\usepackage{stmaryrd}
\usepackage{upgreek}
\usepackage{bm}
\usepackage{cases}
\usepackage{mathtools}

\usepackage{cite}
\usepackage[pagebackref=true,breaklinks=true,letterpaper=true,colorlinks,bookmarks=false,citecolor=red]{hyperref}

\usepackage{algorithm}
\usepackage{algpseudocode}

\usepackage{multirow}
\usepackage{rotating}
\usepackage{booktabs}

\usepackage{enumitem}
\usepackage[olditem,oldenum]{paralist}

\usepackage{alltt}
\usepackage{listings}

\usepackage{url}
\usepackage{xspace}
\usepackage{comment}
\usepackage{color}
\usepackage{afterpage}
\usepackage{pdfpages}
\usepackage{framed}
\usepackage{fancybox}
\usepackage[normalem]{ulem}
\usepackage{dsfont}
\usepackage{soul}

\usepackage{mysymbols}

\begin{document}
\pagestyle{headings}
\mainmatter

\title{Pose from Action: Unsupervised Learning of Pose Features based on Motion} 

\titlerunning{Pose from Action: Unsupervised Learning of Pose Features based on Motion}

\authorrunning{Senthil Purushwalkam, Abhinav Gupta}

\author{Senthil Purushwalkam, Abhinav Gupta}


\institute{Robotics Institute,\\
	Carnegie Mellon University\\
	\email{ \{spurushw@andrew,abhinavg@cs\}.cmu.edu}
}

\maketitle

\begin{abstract}
Human actions are comprised of a sequence of poses. This makes videos of humans a rich and dense source of human poses. 
We propose an unsupervised method to learn pose features from videos that exploits a signal which is complementary to appearance and can be used as supervision: motion. 
The key idea is that humans go through poses in a predictable manner while performing actions. Hence, given two poses, it should be possible to model the motion that caused the change between them.
We represent each of the poses as a feature in a CNN (Appearance ConvNet) and generate a motion encoding from optical flow maps using a separate CNN (Motion ConvNet). 
The data for this task is automatically generated allowing us to train without human supervision.
We demonstrate the strength of the learned representation by finetuning the trained model for Pose Estimation on the FLIC dataset, for static image action recognition on PASCAL and for action recognition in videos on UCF101 and HMDB51.


\end{abstract}

\section{Introduction}
\label{sec:intro}
In recent years, there has been a dramatic change in the field of computer vision. Owing to visual feature learning via convolutional neural networks, we have witnessed major performance gains in different areas including image classification~\cite{russakovsky2014imagenet,szegedy2014going}, object detection~\cite{girshick_arxiv13,girshick2015fast,girshick14CVPR}, 3D scene understanding~\cite{wang2015designing}, pose estimation~\cite{toshev2014deeppose} etc. In most cases, visual features are first learned by training for the classification task on the ImageNet dataset followed by fine-tuning the pre-trained network for the task at hand. 

While this classification based learning framework has yielded significant gains, it is unclear if this is the right approach to visual feature learning. For example, in case of humans, we do not need millions of category-labeled images/videos to learn visual features. Instead, we can learn a visual representation by observing and actively exploring the dynamic world around us. Furthermore, the manual labeling of images remains a significant bottleneck in exploiting a larger number of images to learn visual representations. As a consequence, there has been rising interest in the area of unsupervised feature learning.

\begin{figure}[h!]
	\centering

	\includegraphics[width=\textwidth]{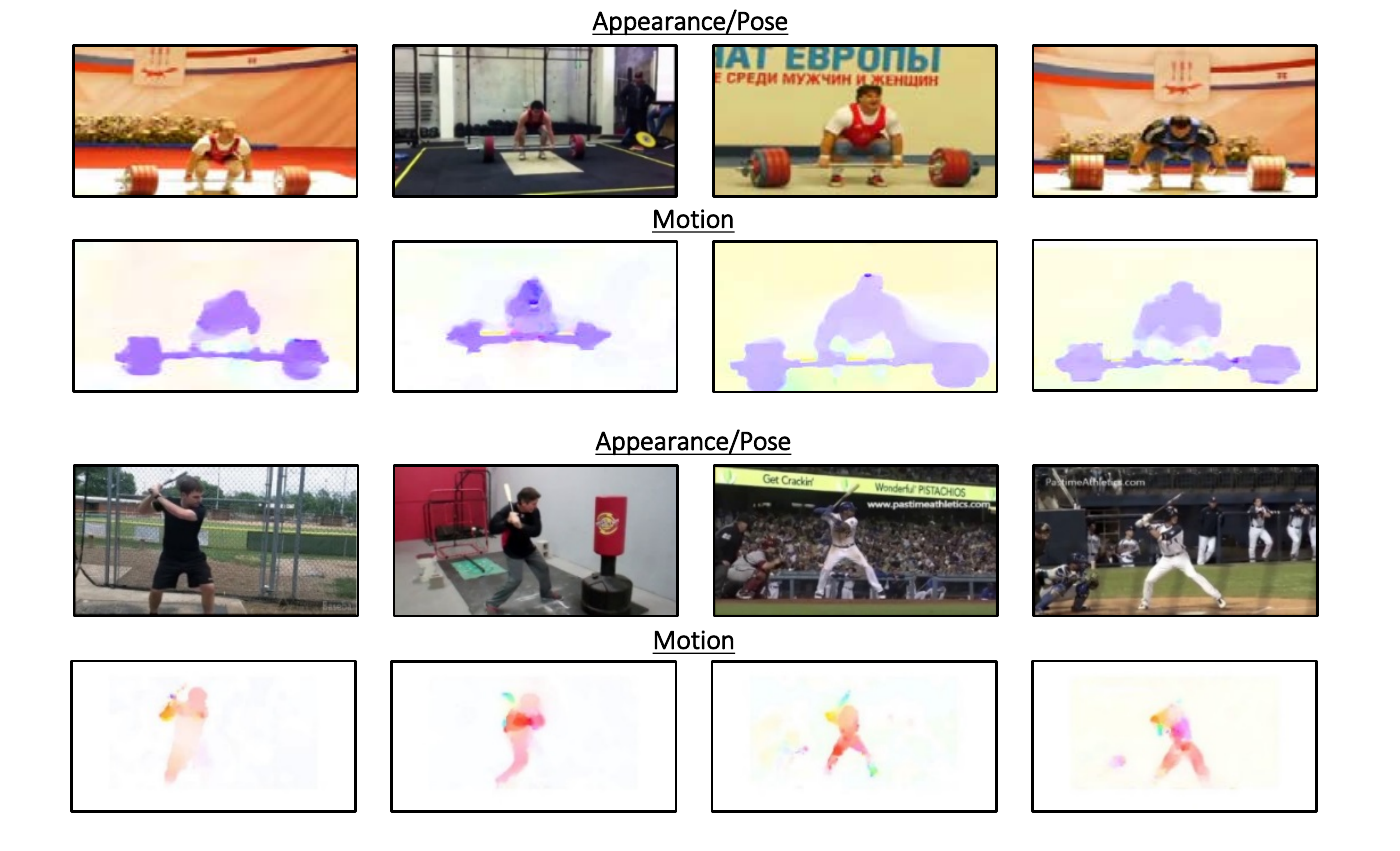}
	\caption{Similar poses are related to similar motions. Hence motion can be used as a supervisory signal to learn appearance representations. We use the following color coding to visualise the optical flow: \protect\capimage{Images/code}}
	\label{fig:teaser}
\end{figure}

There are two paradigms for unsupervised feature learning: generative and discriminative. In the generative learning paradigm, we learn a low-dimensional representation that can be used to generate realistic images. These networks use denoising or reconstruction loss with regularization such as sparsity of the learned space. However, the generative learning approaches have been been not been competitive on tasks like object classification or detection.

In the discriminative learning paradigm, the network is trained using standard back-propagation on an auxiliary task for which ground truth can be easily mined in an automated fashion. 
The hope is that the visual representation learned for this auxiliary task is generalizable and would work for other tasks with simple fine-tuning. 
Owing to the rise of interest in unsupervised learning, many such auxiliary tasks have been proposed in the recent past. 
\cite{doersch2015unsupervised} proposed to take pair of patches sample from an image and predict the relative location of the patches, which seems to generalize to suprisingly well to object detection.
\cite{jayaraman2015learning, agrawal2015learning} proposed an approach to take pair of patches and predict the camera motion that caused the change. The ground-truth for this task is obtained via other sensors which measure ego-motion. 
Finally, ~\cite{wang2015unsupervised} presents an approach to sample a pair of patches via tracking and learn a representation which embeds these patches close in the visual representation space (since they are the same object with some transformations).

While \cite{jayaraman2015learning, agrawal2015learning, wang2015unsupervised} use videos for unsupervised learning, they used other sensors or just different viewpoints to train the appearance models. We argue that there is a complementary and stronger signal in videos to supervise the training of these networks: motion patterns. The key inspiration for our proposed method is that similar pairs of poses are associated with similar motion patterns(See Figure~\ref{fig:teaser}). In this paper, we demonstrate how motion patterns in the videos can act as strong supervision to train an appearance representation. We hypothesize that an appearance representation where poses associated to similar motion patterns cluster together could be useful for tasks like Pose Estimation and Action Recognition.
We believe that the proposed approach is generic and can be used to learn different kinds of pose-encoding appearance representations based on different kinds of videos. Specifically, in this paper, we choose to work with human action videos since the learnt representations can be semantically associated to human poses. We believe that this idea can provide the missing link in unsupervised learning of visual representations for human actions and human poses.

However, there is still one missing link: how do you compare motion patterns. One way is to use  distance metric on hand designed motion features (e.g., 3DHOG, HOF\cite{wang2009evaluation}) or the optical flows maps directly. 
Instead, inspired by the success of the two-stream network\cite{simonyan2014two}, we try to jointly learn convolutional features for both the appearance(RGB) and the motion(optical flow) at the same time. 
Our key idea is to have triplet network where two streams with shared parameters correspond to the first and $n^{th}$ frame in the video; and the third stream looks at $n-1$ optical flow maps. 
All the convolutional streams run in a feedforward manner to produce 4096 dimensional vectors. 
The three streams are then combined to classify if the  RGB frames and optical flow channels correspond to each other \textit{i.e.} does the transformation causes the change in appearance?.
Intuitively, solving this task requires the Appearance ConvNet to identify the visual structures in the frame and encode their poses. The Motion ConvNet is expected to efficiently encode the change in pose that the optical flow block represents. We evaluate our trained appearance network by finetuning on the task of pose estimation on the FLIC dataset\cite{tompson14nips}, static image action recognition on PASCAL VOC\cite{pascal-voc-2010}, and action recognition on UCF101\cite{soomro2012ucf101} and HMDB51\cite{kuehne2011hmdb}. 
We show that these models perform significantly better than training from random initialisation.\\

\section{Related Work}
\label{sec:label}

    \subsubsection*{Unsupervised Learning}
    Training deep learning models in a supervised fashion generally requires a very large labeled training set. This is infeasible and expensive in a lot of cases. This has led to an increase of attention to unsupervised techniques to train these models. Research in unsupervised representation learning can be broadly divided into two categories - generative and discriminative. The approach proposed in this paper belongs to the latter.
    
    Majority of the discriminative approaches involve intelligently formulating a surrogate task which involves learning from an easily available signal. These tasks are designed such that the deep model is forced to learn semantics relevant to us like object labels, human poses, activity labels, etc. 
    In \cite{doersch2015unsupervised}, the formulated task involved predicting the relative location of two patches. Automatically cropping pairs of patches from any image makes the `relative location' signal readily available. The key motivation here is that performing well in this task requires understanding object properties. Hence the Convolutional Neural Network trained to perform this task is shown to perform well on object classification and detection tasks. Similarly, the surrogate task proposed in this paper involves predicting whether a transformation (inferred from optical flow) represents the same transformation as that between a given pair of appearance features.
    
    Unsupervised learning algorithms that learn from videos are extremely valuable since the amount of video data available to us is massive and collecting annotations for them is infeasible. In \cite{wang2015unsupervised}, patches are tracked across frames of videos to generate pairs which are visually dissimilar but semantically same. 
    An unsupervised representation is then learnt by enforcing the similarity on the pair of features extracted for the patches. This structure in the feature space is enforced using a triplet ranking loss which minimises the distance between the pair of features and simultaneously maximises the distance to a feature extracted for a randomly chosen patch.  While this approach shows impressive results on a wide range of tasks, it suffers from two drawbacks. First, the constraint explicitly enforced leads to an appearance representation which is invariant to pose, size and shape changes in an object. Second, the spatially and temporally sparse samples of patches do not make use of all the information available in the videos. In contrast, we attempt to learn a representation that encodes the structural changes by making use of densely sampled pairs of frames to capture a large number of variations in poses.
    
    The unsupervised learning approaches which are closely related to our work are video-based approaches which model similarities or differences across frames\cite{goroshin2015unsupervised,mobahi2009deep,jayaraman2015learning,michalski2014modeling,cadieu2012learning}. A large number of approaches use the idea of temporal coherance to train unsupervised representations. These methods exploit the fact that appearances change slowly between adjacent frames\cite{jayaraman2015slow}. 
    
    A recently proposed approach \cite{jayaraman2015learning} involves learning a representation in which transformations are `predictable'. The feature representation is learnt by specifically enforcing the constraint that similar ego-centric motions should produce similar transformations in the feature space. This approach requires a dataset of video frames annotated with the corresponding ego-poses and hence is not scalable. In our proposed approach, we eliminate this requirement by jointly learning to infer a representation for the transformation from optical flow maps which are easy to compute.

    \subsubsection{Action Recognition and Pose Estimation}The task of recognizing human actions from images and videos has received a lot of attention in computer vision \cite{soomro2012ucf101,kuehne2011hmdb,WangFG15,karpathy2014large}. 
    Activity recognition is a challenging computer vision task since recognizing human actions requires perception of the environment, identifying interaction with objects, \textbf{\textit{understanding pose changes in humans}} and a variety of other sub-problems. 
    Most successful action recognition methods involve using combinations of appearance, pose and motion information as features \cite{yang2010recognizing,delaitre2011learning,maji2011action}.
    A decade of research in action recognition has led to approaches that show impressive performances on benchmark datasets\cite{wang2011action,wang2013action,klaser2008spatio,dollar2005behavior,peng2014action}. The majority of successful algorithms for action classification follow a common pipeline. Appearance or motion features are first extracted either densely or at interest points. This is followed by clustering and generating an encoding. These encoded feature vectors are then classified using various kinds of classifiers.  Recently, deep learning based methods have been extended to action recognition\cite{karpathy2014large}. It has been observed that training deep neural networks directly on stacks of video frames is too computationally expensive and does not lead to significant improvements over handcrafted feature based methods\cite{ji20133d}. More recent methods operate on individual frames independently since it is observed that this gives similar performance as using a stack of frames \cite{karpathy2014large}. The Two-Stream network\cite{simonyan2014two} is a fully-supervised deep-learning based action recognition method which achieves performances comparable to state-of-the-art. It involves training independent spatial and temporal networks whose classification scores are fused to give the final prediction. Deep learning methods have also been extended to estimating poses in images and videos. The task of pose estimation involves estimating the locations of body parts. \cite{toshev2014deeppose} uses a deep neural network based regressor to estimate the coordinates of the parts. The model is recursively applied on patches cropped around the previous prediction to obtain better localisation. In \cite{pfister2015flowing}, a deep convolutional neural network is used to predict heat maps for the location of each body part. The model also uses a spatial fusion technique to capture multi-scale information. 
    
    Actions and Poses are very closely related concepts. An action comprises of a sequence of poses in conjunction with interactions with the environment. Videos are a widely available and rich source of actions. As a consequence, they are also the best source for diverse human poses. In \cite{chen2013watching}, a large collection of unlabelled video is searched to augment training data by finding similar poses using the poselet activation vector\cite{maji2011action}. To the best of our knowledge, the approach proposed in this paper is the first in attempting to learn pose features from videos using deep networks in an unsupervised fashion.
\section{Approach}
\label{sec:approach}

\begin{figure}[t]
    	\centering

    	\includegraphics[width=\textwidth]{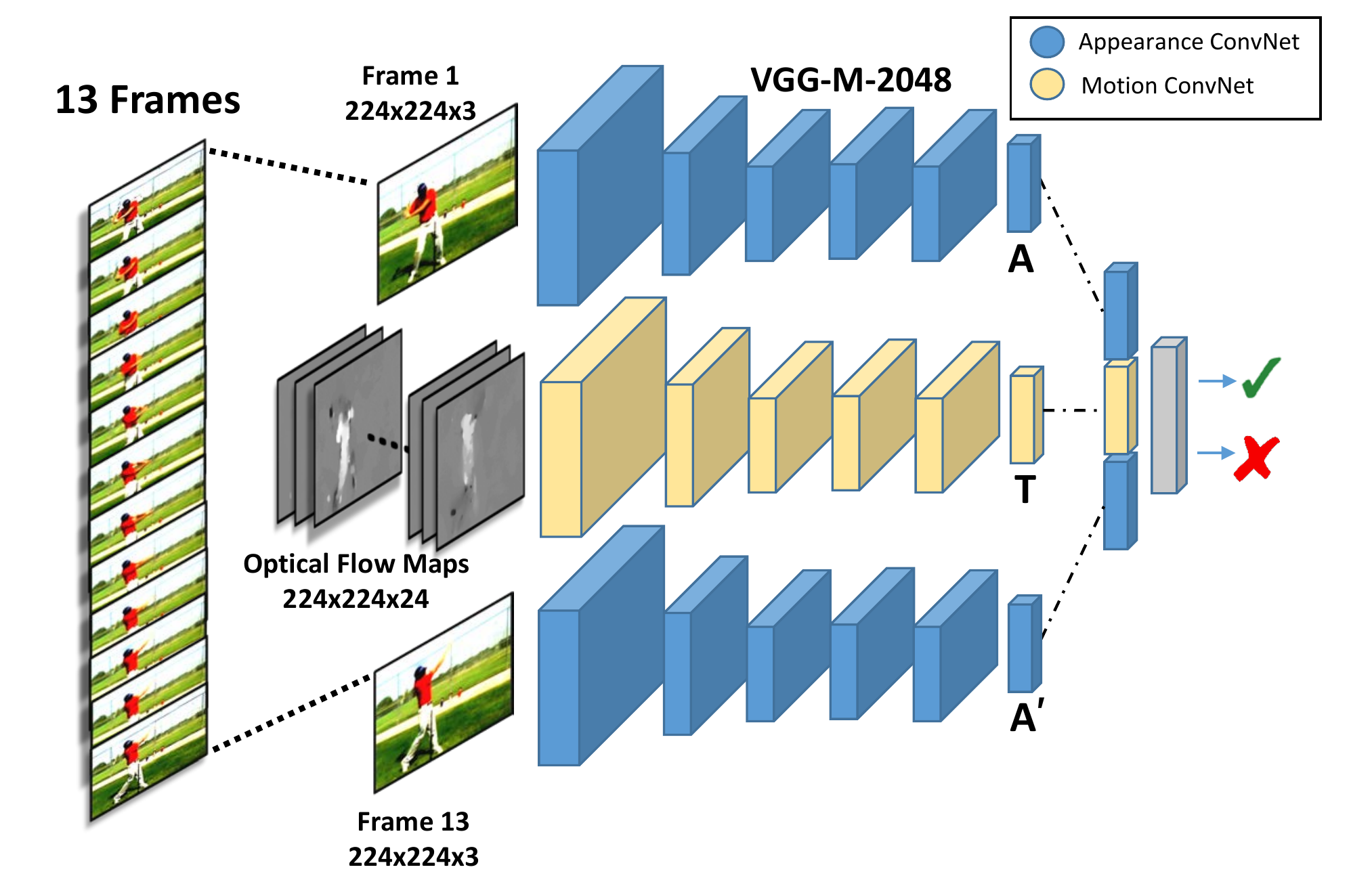}
        \caption{An overview of our approach. Predicting whether a transformation encoding T causes the change in appearance A$\rightarrow$A' requires capturing pose properties.}
        \label{fig:overview}
    \end{figure}

	The goal of this paper is to learn an appearance representation that captures pose properties without the use of any human supervision. We achieve this by formulating a surrogate task for which the ground truth labels are readily available or can be mined automatically. In simple terms, given a change in appearance, the task we formulate involves predicting what transformation causes it. For example, in Figure \ref{fig:overview}, given the appearance of Frame 1 and Frame13, we can predict that the transformation of `swinging the bat' caused the change in appearance. In this section, we first develop an intuitive motivation for the surrogate task and then concretely explain how it can be implemented.
	
    Suppose we want to train a model to predict if a Transformation ${\bf T}$ causes the change in Appearance $\textbf{A}\rightarrow \textbf{A}'$. We would need to have a robust way to encode $\textbf{A}, \textbf{A}'$ and $\textbf{T}$ such that they capture all the information required to solve this task. 
	More specifically, given an image, the appearance representation $\textbf{A}$ needs to localise the object(s) that could undergo a transformation and encode its properties such as shape, size and more importantly, \textbf{\textit{pose}}. 
	On the other hand, given a motion signal (like optical flow, dense trajectories \cite{wang2011action,wang2013action}, etc),  the transformation representation $\textbf{T}$ needs to express a robust encoding that is discriminative in the space of transformations. 
	
	We propose to learn the appearance representation $\textbf{A}$ using a convolutional neural network (Appearance ConvNet in Figure \ref{fig:overview}). We choose to use optical flow maps as the motion signal in our proposed approach. There are a large variety of existing methods like 3dHOG and HOF \cite{chaudhry2009histograms,wang2013action} which can be used to extract an encoding for the optical flow maps. These methods first extract local descriptors in the volume of optical flow maps, and this is generally followed by a bag-of-words model to generate a feature vector. Instead of using these hand-crafted approaches, we propose to jointly learn the motion representation as a Transformation $\textbf{T}$ using a separate convolutional neural network (Motion ConvNet in Figure \ref{fig:overview}). The idea of using two independent networks to represent appearance and motion is very similar to the Two-Stream Network \cite{simonyan2014two} which recently achieved accuracies very close to state-of-the-art in action recognition. The Appearance ConvNet takes as input an RGB image and outputs a feature vector. Similarly, the Motion ConvNet takes as input a stack optical flow maps as input and outputs a feature vector.

	We propose an unsupervised approach to jointly train the Appearance and Motion ConvNets. The key idea of our approach is that given two appearance features ${\bf A}$ and ${\bf A'}$, it should be possible to predict whether a Transformation ${\bf T}$ causes the change ${\bf A \rightarrow  A'}$. This idea is synchronous with \cite{jayaraman2015learning}, where the notion of ego-motions producing predictable transformations is used to learn an unsupervised model.

    Following this intuition, for a video snippet ${\bf i}$, we extract appearance features for Frame $n$ (${\bf A_i(}n{\bf)}$) and Frame $n+\Delta n$ (${\bf A_i(}n+\Delta n{\bf)}$) using the Appearance ConvNet. We then extract motion features for $\Delta n$ optical flow maps for Frames $k$ to $k+\Delta n$ from a random video snippet $j$ (${\bf T_j(}k,k+\Delta n)$) using the Motion ConvNet. 
    We then use two fully connected layers on top of the three concatenated features to predict whether the transformation ${\bf T_j(}k,k+\Delta n)$ could cause the change ${\bf A_i(}n{\bf) \rightarrow  A_i(}n+\Delta n)$ $i.e.$ $${\bf T_j(}k,k+\Delta n) = {\bf T_i(}n,n+\Delta n)$$
    We randomly (and automatically) sample $i$,$n$,$j$, $k$ and keep $\Delta n$ fixed. This makes the positive and negative labels readily available \textit{i.e.} the positive examples are the triplet samples where $i=j$ and $n=k$. All the others samples could be treated as negatives, but to account for videos with repetitive actions (like walking), we mine negatives from other videos \textit{i.e.} we do not use samples where $i=j$ and $n\ne k$. Fixing $\Delta n$ to a constant value is necessary since we need to fix the filter size in the first layer of the Motion ConvNet.
    
    In summary, the joint unsupervised learning pipeline consists of one Motion ConvNet, two instances of the Appearance ConvNet and a two-layer fully connected neural network on top. The parameters of the two Appearance ConvNets are shared since we expect both networks to encode similar properties. Overall the joint system of three neural networks can be treated as one large neural network. This allows us to use standard back propagation to train all the components simultaneously.

    \subsection*{Implementation Details}
	In our experiments, we fix $\Delta n = 12$ \textbf{i.e.} we sample pairs of frames which are separated by 12 frames. We follow the VGG-M architecture for the Appearance ConvNet and Motion ConvNet till the FC6 layer. The only difference is the size of Conv1 filters in the Motion ConvNet which has 24 channels instead of 3 to accommodate convolution on 24 optical flow maps (12 in the $x$-direction and 12 in the $y$ direction). This gives us a 4096-dimensional vector representation for each of $\textbf{A}$, $\textbf{A}'$ and $\textbf{T}$. We then concatenate the three feature vectors to get a 12288 dimensional vector and use a fully connected neural network to perform the binary classification. The first fully-connected layer has 4096 output neurons followed by second fully connected layer with 2 output neurons. A softmax classifier is then used to predict the binary labels. 

	\subsubsection*{Patch and Optical Flow Mining} In order to train the binary classification model, we require a large collection of pairs of frames, the correct block of optical flow maps between them and multiple negative samples of optical flow blocks. As the training set, we use a large collection of video which contain humans performing actions. This set is formed by combining the training set videos from the UCF101 \cite{soomro2012ucf101}(split1), HMDB51 \cite{kuehne2011hmdb} (split1) and the ACT\cite{WangFG15} datasets. 
    For every pair of consecutive frames we precompute the horizontal and vertical directional optical flow maps using the OpenCV GPU implementation of the TVL1 algorithm\cite{marzat2009real}. 

    As inputs to the Appearance ConvNet we randomly sample a spatial location and crop 224x224 patches at that location from two frames separated by $\Delta n(=12)$ frames. For the Motion ConvNet, we sample the $224$x$224$ patches from each of the 12 horizontal and 12 vertical flow maps in between the two sampled frames at the same location, as the positive (label$=1$) which gives us a 224x224x24 dimensional array. As the negative examples (label$=0$), we randomly sample another 224x224x24 block from a random spatial location in a randomly picked video. During training, we pick the negatives from the same batch in the mini-batch stochastic gradient descent procedure and ensure that negative flow blocks are not picked from the same video as the appearance frames.
    We also augment the training data by randomly applying a horizontal flip on a (Frame $n$, Frame $n+\Delta n$, Optical Flow Block) triplet. Since all motion signals also make sense in the reverse direction temporally (they do not necessarily hold any semantic value), we also randomly reverse some triplets \textit{i.e.} (Frame $n+\Delta n$, Frame $n$, reversed optical flow block).
    
    For the joint training procedure, we use a batchsize of 128 $i.e.$ 128 pairs of patches. The SoftMax Loss is used to compute the errors to train the network. We initially set the learning rate to $10^{-3}$, momentum to $0.9$ and train for 75,000 iterations. We then reduce the learning rate to $10^{-4}$ and train for 25,000 iterations. At convergence, the joint system performs around ~96\% on the formulated binary classification task for a held out validation set (note that the baseline performance is 66\%  since we have two negatives for each positive).
    
\section{Experiments}
\label{sec:experiments}

    The efficacy of unsupervised feature representation learning methods are generally tested on tasks for which the learnt representation might prove useful. First, the learnt representations are finetuned for the task using either the full labelled dataset (generally trained for a small number of iterations) or a small subset of the training set. Then these finetuned models are tested to provide evidence for the transferable nature of the representation learnt.
    
    We follow a similar strategy and perform an extensive evaluation of our unsupervised model to investigate the transferability of the learned features. In order to build a better understanding of what the models learn, we first perform a qualitative analysis of the models. As explained before, since our unsupervised model is trained on action videos, this leads to an appearance representation (Appearance ConvNet) that is expected to capture pose properties well. 
    Feature representations that capture pose properties are valuable for estimating human poses.
    Another domain where pose information proves immensely useful \cite{chen2013watching,yang2010recognizing,delaitre2011learning,maji2011action} is recognizing human actions since any action involves a series of poses. Following this intuition, we test our learned representation for the Pose Estimation and Action Recognition tasks.
    
    We also compare our method to two popular and recent unsupervised representation learning methods which also attempt to learn from videos. The results demonstrate the superiority of our learnt representation for these tasks. The first unsupervised model, proposed by Wang et. al in \cite{wang2015unsupervised}, involves enforcing the constraint that two transformed versions of the same object (different viewpoint, pose, size, etc) needs to represent the same point in the feature space. This leads to a feature representation that is invariant to pose, shape and size changes. The second model, proposed in \cite{jayaraman2015slow}, involves enforcing temporal coherence in the feature space by imposing a prior on the higher order derivatives to be small. This is trained jointly with the classification loss for the supervised task. We compare to this model since it is the most recently introduced unsupervised technique for videos.
    

\subsection{Qualitative analysis of learned models}
\begin{figure}[h!]
    	\centering
    	\includegraphics[width=\textwidth]{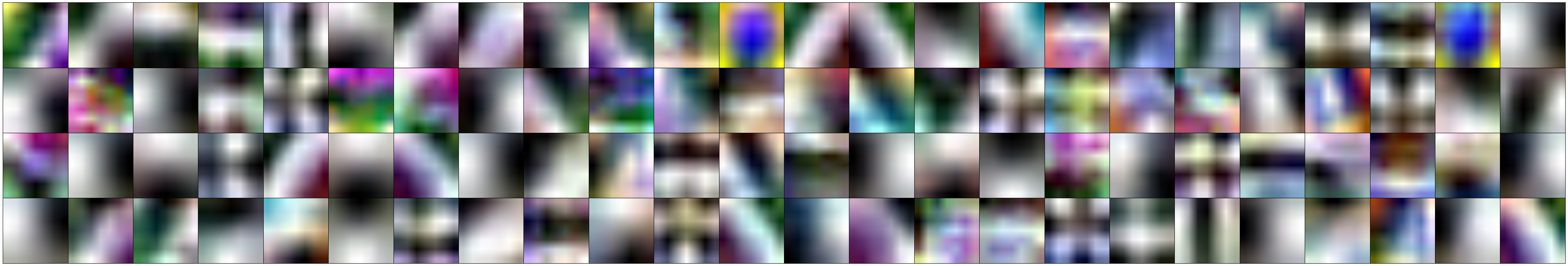}
    	\caption{Visualisations of filters in the first convolution layer in the Appearance ConvNet.}
    	\label{fig:visconv1}
    \end{figure}

    The first layer of a convolutional neural network is often visualised to verify that the network learns meaningful representations. We present the visualisations of the 96 filters in the first convolutional layer of the Appearance ConvNets in Figure \ref{fig:visconv1}. Clearly, the visualisation shows that the filters learn to model gradient like features. 
    
\begin{figure}[h!]
    	\centering
    	
    	\includegraphics[width=0.85\textwidth]{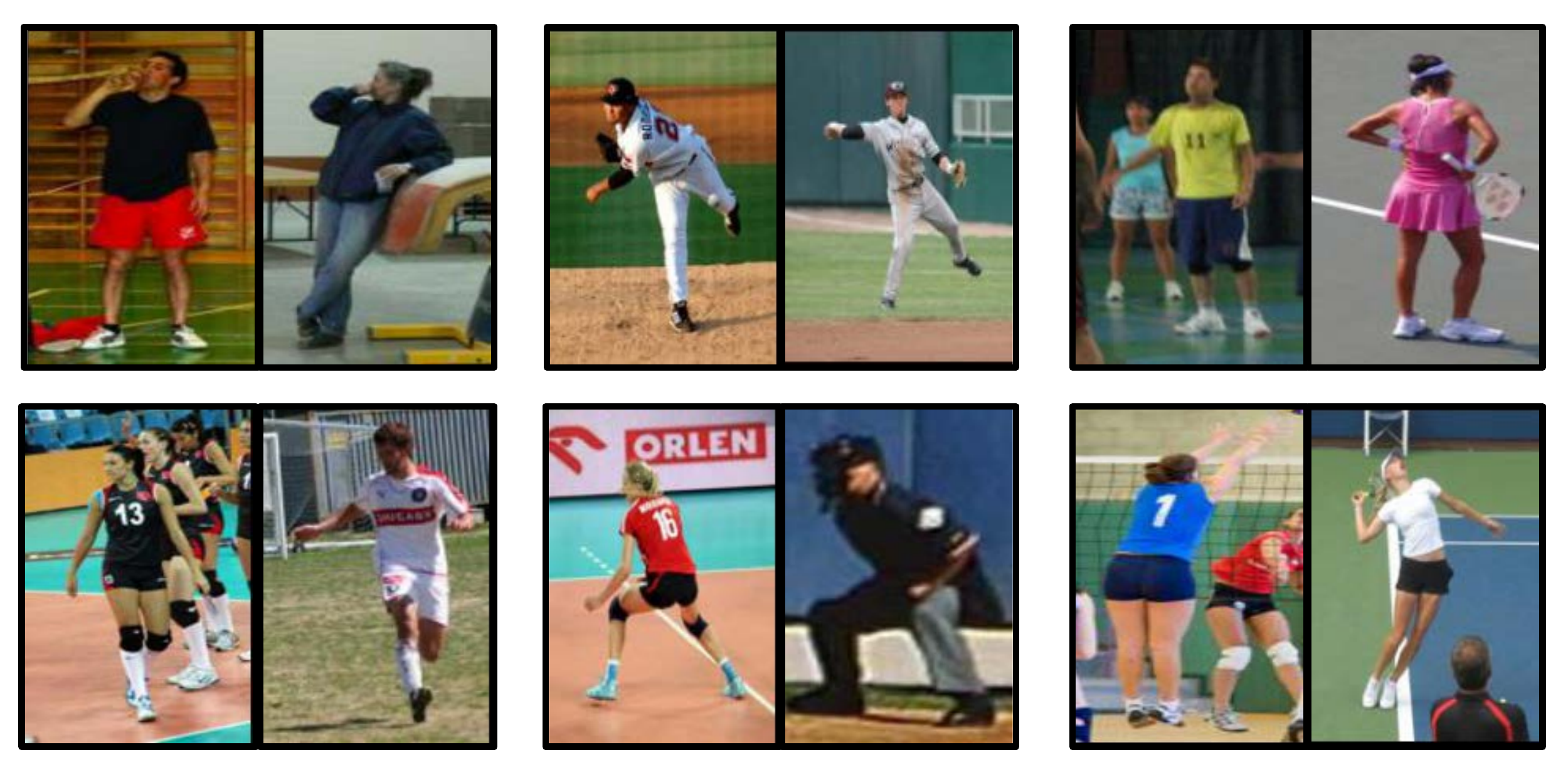}
    	\caption{Closest image pairs in the FC6 feature space of the Appearance ConvNet.}
    	\label{fig:nearest}
    	\vspace{-15pt}
    \end{figure}

    We investigate the pose capturing capability of the learned unsupervised representation in the Appearance ConvNet by visualising closest pairs in the FC6 feature space. We first compute the appearance features for all image in the Leeds Sports Pose(LSP) dataset \cite{Johnson10}. We randomly sample images and find the closest image in the rest of the dataset use the Euclidean distance between the appearance features extracted. We present these closest pairs in Figure \ref{fig:nearest}. From these pairs, it is evident that the Appearance ConvNet is able to match poses reasonably well. This observation suggests that the Appearance ConvNet indeed attempts to capture the pose properties of humans. 
    
\subsection{Pose Estimation}
    The task of estimating human poses from videos and images is an important problem to be solved and has received widespread attention. 
    In its most simple form, the task is defined as correctly localising the joints of the human. Computer vision research focusing on pose estimation has given rise to a large number benchmarks which contain videos and images \cite{tompson14nips,andriluka20142d,pfister2015flowing} with their annotated joints. We evaluate the efficacy of our learnt Appearance ConvNet by testing it for estimating human poses in the Frames Labelled in Cinema (FLIC) dataset \cite{tompson14nips}. This dataset contains 5003 images with the annotated joints collected using crowd-sourcing. The train and test splits contain 3987 and 1016 images respectively.
        
\begin{figure}[h!]
    	\centering
    	\includegraphics[width=\textwidth]{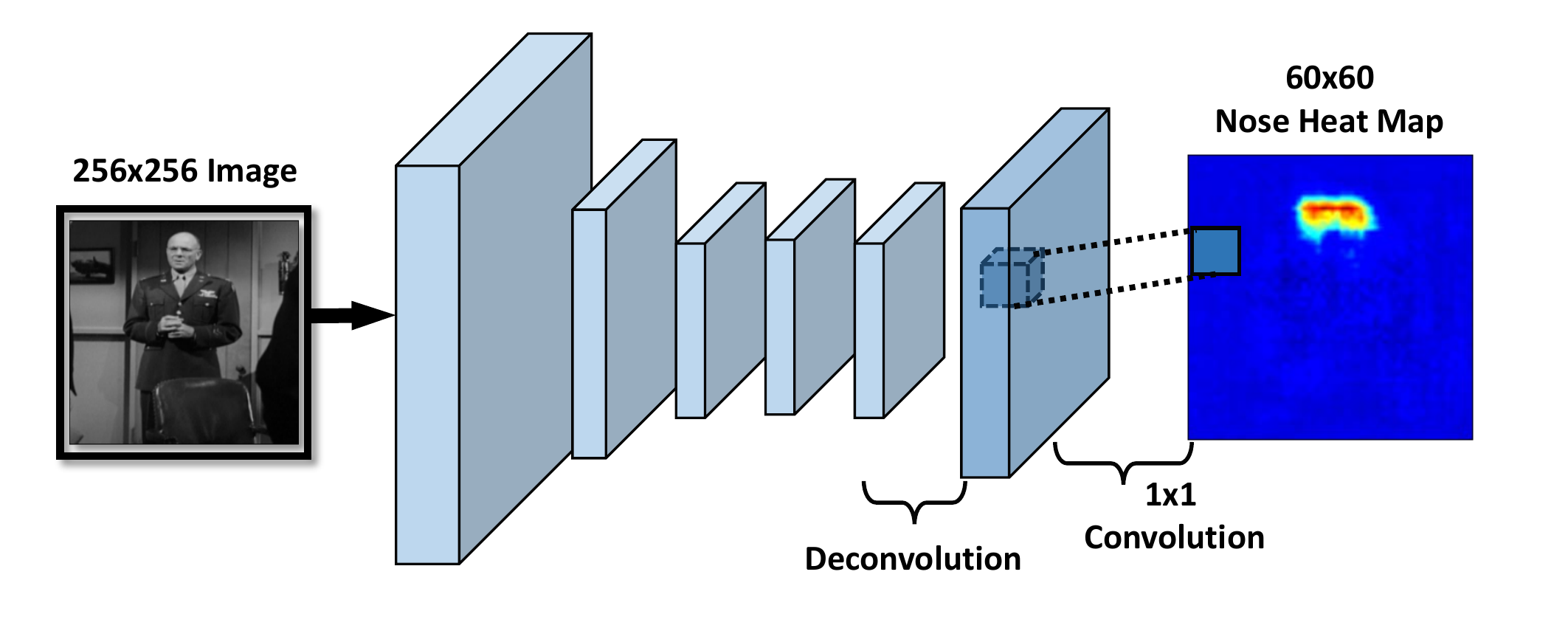}
    	\caption{Architecture of the pose estimation network. First 5 layers copied from VGG-M, followed by a deconvolution layer. A 1x1 convolution layer is then used to predict each output heat map. }
    	\label{fig:posenet}
    \end{figure}

    We design a simple deep learning based pose estimation architecture to allow us the freedom to accommodate other unsupervised models. This also improves interpretability of the results by minimising the interference of complementary factors on the performance. Figure \ref{fig:posenet} presents an overview of the architecture we use to perform pose estimation (referred as Pose ConvNet). We copy the VGG-M\cite{chatfield2014return} architecture till the fifth convolution layer (Conv5). This is followed by a deconvolution layer to upscale the feature maps. 
    Then 1x1 convolutions are used to predict heat maps for each body point to be estimated. This network architecture is partly inspired from \cite{tompson2015efficient}. The predicted heat maps are 60x60 dimensional. The FLIC dataset contains annotations for the $(x,y)$ coordinates of 9 points on the body (nose, shoulders, elbows, hips and wrists). Hence our architecture uses nine separate 1x1 convolutional filters in the last layer to predict the heat maps for each annotated point.
    
    \subsubsection{Preprocessing} Since the task we are evaluating is pose estimation (and not detection), we first need to crop the images around the annotated human. We do this by expanding the torso ground truth box by a fixed scale on all images. We then rescale all cropped images to 256x256. For each of the new cropped and rescaled images, we generate nine 60x60 ground truth heat maps, one for each of the joints. The heat map values are scaled between [-1,1] such that -1 represents background and +1 represents the presence of a body part. These ground truth heat maps are used to train the convolutional neural network. Since each ground truth heat map has only one positively activated pixel, the data is not sufficient to train the whole neural network. So we augment the data by activating a 3x3 neighbourhood in the heat maps.
    
    \subsubsection{Training} We use the Euclidean loss to compute the error signal for each output heat map in the Pose ConvNet. Since we have 9 ground truth heat maps, we have access to 9 error signals. We use standard backpropagation to train the network and average the gradients from all the nine euclidean losses. 
    Training the Pose ConvNet directly using this procedure converges at predicting all pixels as -1 in the heat maps since the number of positive pixels are still very small in the ground truth.
    In order to overcome this, we reweigh the gradient w.r.t. a positive ground truth pixel by the inverse of number of total number of positive pixels and similarly for the negative pixels. This ensures that the sum of gradients for the positive pixels is equal to the sum of gradients for the negative pixels.
    
    \subsubsection{Evaluation} The trained Pose ConvNet maps are used to generate body part heat maps for each of the test images in the FLIC dataset. The highest scoring 20 pixels are identified in each heat map and the location of the centroid of these pixels is used as the prediction for that body part. Various evaluation metrics have been studied in the past for evaluating pose estimations methods \cite{chen2014articulated,pishchulin2012articulated,toshev2014deeppose}. We report accuracies using the Strict Percentage of Correct Parts(PCP) and the Percentage of Detected Joints (PDJ) metrics. We use the code made available by \cite{chen2014articulated} to compute these metrics.
    
    We train four models using the Pose ConvNet architecture to investigate the strength and transferability of our unsupervised representation. We test our unsupervised Appearance ConvNet by copying parameters to the first five convolutional layers of the Pose ConvNet and randomly initialising the last two layers. 
    We then finetune the model on the training data from the FLIC dataset. We follow a similar procedure for the baseline model \cite{wang2015unsupervised}. We also train an instance of the Pose ConvNet from scratch with random initialisation to compare with our model. The Strict PCP accuracies for these models are presented in Table \ref{tab:strictpcp} and the PDJ accuracies at varying precision values is presented in Table \ref{tab:pdj}. The Appearance ConvNet beats the accuracy of the randomly initialised baseline by a large margin indicating that the Appearance ConvNet indeed learns a representation useful for Pose Estimation. We also observe a significant increase over the baseline unsupervised model \cite{wang2015unsupervised} suggesting that the representation learnt by the Appearance ConvNet encodes properties not captured in the baseline.
    Surprisingly, we observe that when the Pose ConvNet is initialised with a model trained to perform action classification on the UCF101 dataset, it performs worse than random initialisation. This suggests the invariances learned due to semantic action supervision are not the right invariances for pose-estimation. Therefore, using an unsupervised model leads to unbiased and stronger results. In our experiments, we also observe that using Batch Normalization\cite{ioffe2015batch} while training the Pose ConvNet initialised with Appearance ConvNet leads to a very narrow increase in performance (~1.5\% in PCP).

    \begin{table}[h]
    \begin{center}
    \caption{Results for the Strict PCP Evaluation for Pose Estimation on the FLIC Dataset}
    \vspace{-8pt}
    \label{tab:strictpcp}
    \setlength{\tabcolsep}{10.5pt}
    \begin{tabular}{@{} l c c @{}}
    \toprule
        &  \multicolumn{2}{c}{\textbf{Body Part}} \\
    \textbf{Initialisation}    &  \textbf{Upper Arms}  &  \textbf{Lower Arms}\\
    \midrule
    \midrule
    Random   & 51.9 & 19.3 \\
    Wang et. al Unsupervised\cite{wang2015unsupervised}  &  52.8 & 19.7\\
    UCF101 Action Classification Pretrained  &  46.7 & 17.8\\
    Ours   & \textbf{57.1} & \textbf{24.4} \\
    \midrule
    ImageNet Classification Pretrained&  65.6 & 34.3 \\ 
    
    \bottomrule
    \vspace{-50pt}
    \end{tabular}
    \end{center}
    \end{table}

    \begin{table}[h]
    \begin{center}
    \caption{Results for the PDJ Evaluation for Pose Estimation on the FLIC Dataset}
    \vspace{-8pt}
    \label{tab:pdj}
    \setlength{\tabcolsep}{3.5pt}
    \begin{tabular}{@{} l c c c c | c c c c @{}}
    \toprule
        &  \multicolumn{4}{c}{Elbow} & \multicolumn{4}{c}{Wrist} \\
        \cline{2-9}
    Initialisation ~~~~~~~~~Precision$\rightarrow$   &  0.1 & 0.2 & 0.3 & 0.4 & 0.1 & 0.2 & 0.3 & 0.4\\
    \midrule
    \midrule
    Random   & 20.0 & 47.3 & 63.9 & 74.8 & 17.2 & 36.1 & 49.6 & 60.8 \\
    UCF101 Pretrained  &  18.5 & 44.8 & 61.0 & 71.1 & 16.5 & 34.8 & 45.2 & 53.2\\
    Wang et. al Unsupervised\cite{wang2015unsupervised}  &  23.0 & 48.3 & 66.5 & \textbf{77.6} & 19.1 & 36.6 & 46.7 & 55.1\\
    Ours   & \textbf{28.0} & \textbf{54.6} & \textbf{68.8} & \textbf{77.6} & \textbf{20.1} & \textbf{40.0} & \textbf{51.6} & \textbf{60.8} \\
    \midrule
    ImageNet Pretrained&  34.8 & 62.0 & 74.7 & 82.1 & 29.0 & 48.5 & 59.3 & 66.7 \\ 
    
    \bottomrule
    \vspace{-20pt}
    \end{tabular}
    \end{center}
    \end{table}

\subsection{Action Recognition}
    For the task of action recognition, we use the UCF101 and HMDB51 datasets. We test on split1 for both datasets since we use the same split to train our unsupervised models. UCF101 consists of 9537 train and 3783 test videos, each of which shows one of 101 actions. 
    The HMDB51 dataset is a considerably smaller dataset which contains 3570 train and 1530 test videos and 51 possible actions. Due to the size of the HMDB51 dataset, overfitting issues are accentuated. Therefore, training deep models from scratch on this dataset is extremely difficult. In \cite{simonyan2014two}, the authors suggest multiple data augmentation techniques to alleviate these issues.
    In our experiments, we witnessed that initialising from our unsupervised model also helps in overcoming this issue to a certain extent which is reflected in the results. We also compare our results to \cite{wang2015unsupervised} as before. 
    
    Similar to the Pose ConvNet, we use the Appearance ConvNet as an initialisation for action recognition to investigate its performance. We use the same architecture as the Appearance ConvNet(VGG-M till FC6) followed by two randomly initialised fully-connected layers at the end to perform classification. 
    The first fully-connected layer has 2048 output neurons, and the second fully-connected has 101 output neurons for classification on UCF101 and 51 output neurons for classification on HMDB51.The softmax classification loss is used to train the action classification network. The input to the network is a random 224x224 crop from any frame in the video. During training, we use a batch size of 256, which gives us 256 crops of dimension 224x224 sampled from random videos.
    After intialising with the appropriate parameters, we train the whole model for 14k iterations using learning rate as $10^{-3}$ and for another 6k iterations using learning rate as $10^{-4}$. 
    
    \subsubsection{UCF101 And HMDB51} For testing the network, we uniformly sample 25 frames from the test video. From each of the 25 frames, we sample 224x224 crops from the corners and the center. We also generate flipped versions of each of these samples giving us 250 samples per video. We compute the predictions for each of the samples and average them across all samples for a video to get the final prediction. The classification accuracies on both datasets are reported in Table \ref{tab:spatial}.
    We also present the results achieved by \cite{simonyan2014two} for training from scratch and training from a network pretrained on ImageNet for classification. 
    The results reflect improvement over training from random initialisation by significant margins - 12.3\% on UCF101 and 7.2\% on HMDB51. This clearly indicates that the Appearance ConvNet encodes transferable appearance features which are also useful for action recognition.
    Surprisingly, finetuning just the last 2 fully connected layers also beats training from scratch on HMDB51 and scores comparably on the UCF101 dataset. This further emphasises the transferable nature of the Appearance ConvNet.

    \begin{table}[t]
    \begin{center}
    \caption{Results for the Appearance Based action recognition on UCF101 and HMDB51}
    \vspace{-8pt}
    \label{tab:spatial}
    \setlength{\tabcolsep}{4.5pt}
    \begin{tabular}{@{} l c  c    c  @{}}
    \toprule
        &  & \multicolumn{2}{c}{Dataset} \\
    Initialisation    &  Finetuning/Training  & UCF101  &  HMDB51\\
    \midrule
    \midrule
    Random & Full Network  & 42.5\% & 15.1\% \\
    Wang et. al Unsupervised\cite{wang2015unsupervised} & Full Network &  41.5\% & 16.9\%\\
    Ours & Full Network  & {\bf 55.4\%} & {\bf 23.6\%} \\
    Ours & Last 2 layers  &  41.4\% &  19.1\%\\
    \midrule
    ImageNet & Full Network  &  70.8\% & 40.5\% \\ 
    
    \bottomrule
    \end{tabular}
    \end{center}
    \end{table}

    \subsection{Static Image PASCAL Action Classification} 
    For the second baseline model~\cite{jayaraman2015slow}, classification accuracies are reported on the Pascal Action Classification dataset. The task involves classifying static images into one of the 10 action classes. The experiment used in \cite{jayaraman2015slow}, involves training the model using just 50 randomly sampled training images while simultaneously enforcing the prior they formulate. 
    To allow fair comparison, we finetune our Appearance ConvNet using 50 randomly sampled images. We train an action classification network similar to the network described above but with 10 output neurons. The results for this experiment are reported in Table \ref{tab:pascal}. The Appearance ConvNet shows an improvement of 2.5\% over ~\cite{jayaraman2015slow} on this task.

    \begin{table}[t]
    \begin{center}
    \caption{Results for action recognition accuracy in static images using just 50 randomly sampled training images from PASCAL VOC2010 dataset (mean over 5 runs)}
    \vspace{-8pt}
    \label{tab:pascal}
    \setlength{\tabcolsep}{15.5pt}
    \begin{tabular}{@{} l  c    c  @{}}
    \toprule
    Method  & Accuracy\\
    \midrule
    \midrule
    Random Initialisation (taken from \cite{jayaraman2015slow}) \ & 15.34\%\\
    SSFA\cite{jayaraman2015slow}  & 20.95\%\\
    Appearance ConvNet initialisation &  \textbf{22.7\%}\\
    \bottomrule
    \vspace{-30pt}
    \end{tabular}
    \end{center}
    \end{table}
\section{Conclusion and Future Work}
\label{sec:conclusion}

    In this paper, we present an unsupervised algorithm that takes advantage as the motion signal in videos as supervision to train an appearance representation. We train the unsupervised system on action videos in order to force the appearance representation to learn pose features. We demonstrate this property of the feature representation using qualitative results and quantitative results on Pose Estimation in the FLIC dataset, Action Recognition in videos on the UCF101 and HMDB51 datasets and still image action recognition on PASCAL VOC. The finetuning results emphasise the highly transferable nature of the representations learned. We compare to two other video-based unsupervised algorithms and show that our trained representation performs better consistently on these tasks. As a future goal, an interesting direction to pursue would be extending this method to generic videos. 
\section{Acknowledgement}
\label{sec:ack}

Supported by the Intelligence Advanced Research Projects Activity (IARPA) via Department of Interior/ Interior Business Center (DoI/IBC) contract number D16PC00007. The U.S. Government is authorized to reproduce and distribute reprints for Governmental purposes notwithstanding any copyright annotation thereon. Disclaimer: The views and conclusions contained herein are those of the authors and should not be interpreted as necessarily representing the official policies or endorsements, either expressed or implied, of IARPA, DoI/IBC, or the U.S. Government.


\bibliographystyle{splncs}
\bibliography{senthil}
\end{document}